\pdfoutput=1 

\documentclass[a4paper,conference]{IEEEtran}
\usepackage{graphicx}
\usepackage{xcolor}
\usepackage{tikz}

\usepackage{pgfplots}
\usepackage{caption}
\usepackage{subcaption}
\usepackage{float}
\usepackage{cite}

\ifCLASSINFOpdf
\else
\fi
\usepackage{amsmath}
\usepackage{url}
\usepackage{hyperref}

\begin{document}
%
\title{\vspace{-9pt}Towards Individual Grevy's Zebra Identification\\via Deep 3D Fitting and Metric Learning\vspace{-80pt}}

\begin{small}
\author{\vspace{60pt}\\ \ \IEEEauthorblockN{Maria Stennett}
\IEEEauthorblockA{University of Bristol, UK\\
gh18931@bristol.ac.uk}
\and
\vspace{60pt}\\ \ \IEEEauthorblockN{Daniel I. Rubenstein}
\IEEEauthorblockA{Princeton University, USA\\
dir@princeton.edu}
\and
\vspace{60pt}\\ \ \IEEEauthorblockN{Tilo Burghardt}
\IEEEauthorblockA{University of Bristol, UK\\
tilo@cs.bris.ac.uk }
}
\end{small}


%


\maketitle

\begin{figure}
\includegraphics[width=250pt,height=0pt]{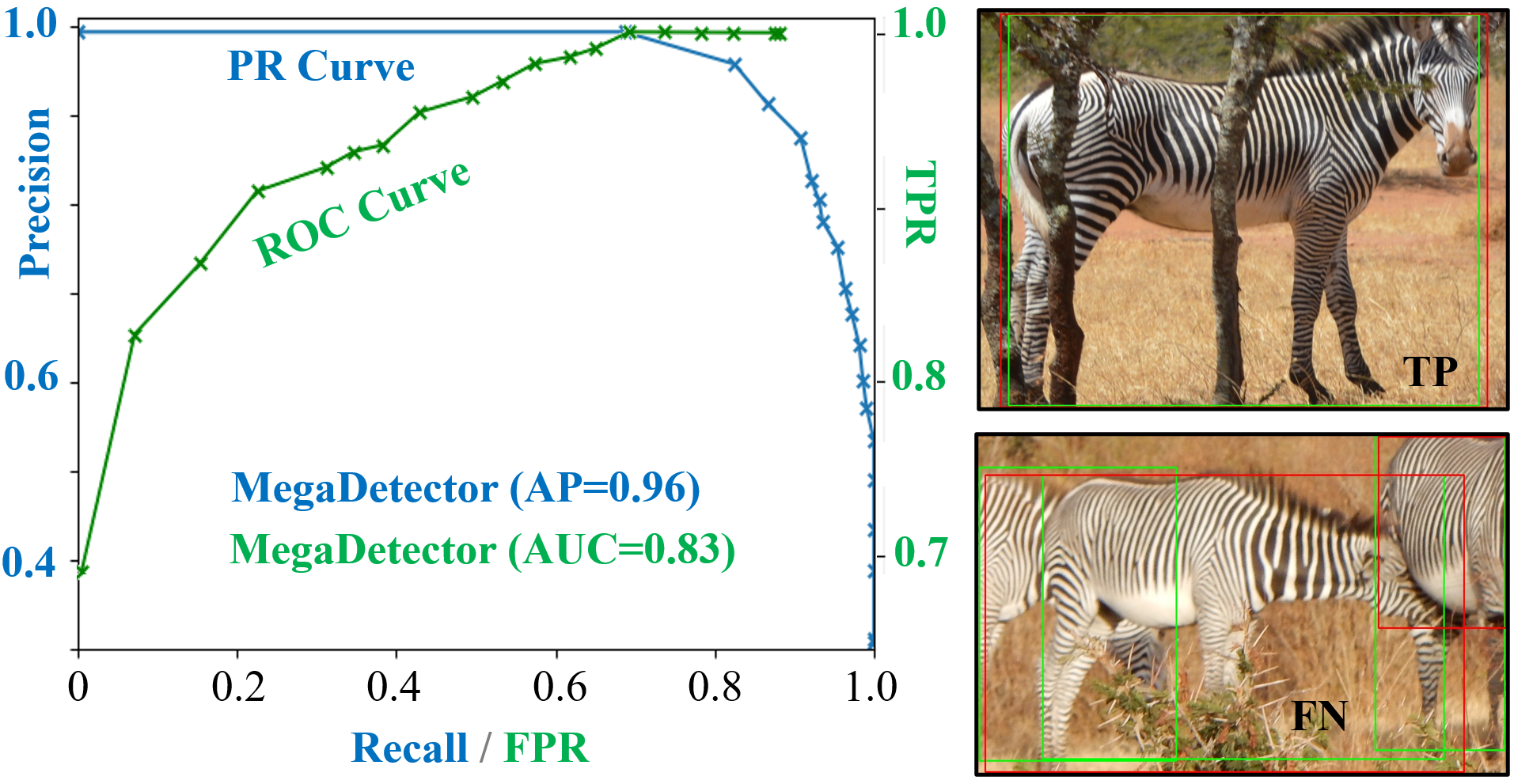}
\vspace{-60pt}
\end{figure}
\textbf{\textit{Abstract -- }This paper combines deep learning techniques for species detection, 3D model fitting, and metric learning in one pipeline to perform individual animal identification from photographs by exploiting unique coat patterns at below real-time. This is the first work to attempt this and, compared to traditional 2D bounding box or segmentation based CNN identification pipelines, the approach provides effective and explicit view-point normalisation and allows for a straight forward visualisation of the learned biometric population space. Note that due to the use of metric learning the pipeline is also readily applicable to open set and zero shot re-identification scenarios. We apply the proposed approach to individual Grevy's zebra \textit{(Equus grevyi)} identification and show in a small study on the SMALST dataset that the use of 3D model fitting can indeed benefit performance. In particular, back-projected textures from 3D fitted models improve identification accuracy from 48.0\% to 56.8\% compared to 2D bounding box approaches for the dataset. Whilst the study is far too small accurately to estimate the full performance potential achievable in larger-scale real-world application settings and in comparisons against polished tools, our work lays the conceptual and practical foundations for a next step in animal biometrics towards deep metric learning driven, fully 3D-aware animal identification in open population settings. We publish network weights and relevant facilitating source code with this paper for full reproducibility and as inspiration for further research.}


%
\IEEEpeerreviewmaketitle

\section{Introduction and Biological Context}
\begin{figure}[t]
    \centering
    \vspace{-25pt}\includegraphics[width=255pt,height=320pt]{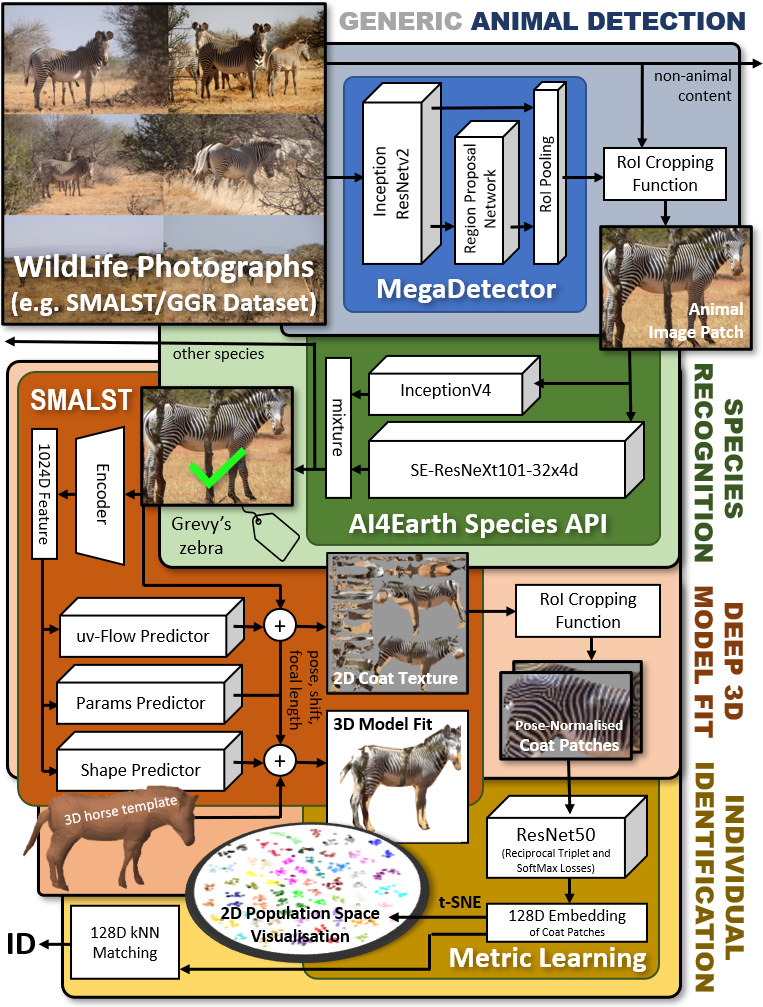}
    \caption{\begin{footnotesize}\textbf{System Overview.} We integrate four distinct deep learning systems into a fully automated pipeline for animal identification: \textbf{\textit{(blue)}}~animal detection and localisation via the MegaDetector~\cite{MegaDetector}; \textbf{\textit{(green)}}~species recognition/confirmation via the Microsoft AI for Earth species classification API~\cite{sc_model}; \textbf{\textit{(orange)}}~3D zebra model fitting via the SMALST framework~\cite{ZuffiICCV2019} and subsequent model back-projection of the visible texture portion of the coat pattern leading to a normalised surface texture; \textbf{\textit{(yellow)}}~Extracted textures are used to learn a latent population space via ResNet-driven metric learning~\cite{andrew2020visual}. Novel images projected into this space can be classified by k-nearest neighbour techniques, resulting in zebra identities. \end{footnotesize}}\vspace{-16pt}
    \label{fig:pipeline_overview}
\end{figure}

Timely population data is a quintessential pillar required for most ecological studies and, critically, for effective conservation efforts~\cite{tuia}. In the past, accurate data on where and when individuals were seen was limited to only a few animals who were fitted with GPS tags or to those whose distinctive features easily stood out. Thus, inferences about population size, structure of dynamics associated with changes over time in population numbers, individual associations or habitats were limited because they were based on small samples. Biometric identification from photographs based on unique coat patterns avoids tagging and its computerisation addresses scaling issues. For plains zebras, for instance, symbolic stripe codes on manual filing cards were described by Klingel and later Petersen~\cite{petersen} in the mid-20th century. Computerised photographic identification started in the 1990s with first approaches by Hiby and Lovell~\cite{hiby}, which explicitly normalised for viewpoint using semi-manual model fitting. Other early ID methods either relied on user input or were limited to high quality~\cite{lahiri,sherley} or pre-processed imagery. The arrival of local descriptors~\cite{hotspotter} in combination with deep learning approaches~\cite{parham} changed the field. Routine identification of individuals based on unique visual appearance of many species is now possible with good success for reasonable quality imagery~\cite{berger,tuia}. However, animal pose variations and deformations still cause challenges in individual re-identification despite the fact that metric learning~\cite{andrew2020visual} has opened new horizons for performance~\cite{schneider}, open set applicability and population space visualisation~\cite{andrew2020visual}. Today, no framework exists that unifies modern approaches into an explicit 3D-aware deep metric learning setting. All existing automated animal ID techniques still rely on the implicit encoding of pose and viewpoint invariance within the classification framework or descriptor despite the availability of first explicit deep 3D fitting techniques for animals~\cite{ZuffiICCV2019}. In response, we propose a first deep learning ID approach that integrates all desired aspects in a single pipeline~(see Fig.~\ref{fig:pipeline_overview}). It is available at 
\url{https://github.com/Lm0079/grevys-zebra-individual-identification}.

\section{Datasets}

We use the 148 Grevy’s zebra (Equus grevyi) images of the SMALST repository~\cite{ZuffiICCV2019} for full system evaluation and combine it with 687 images from the WCS dataset~\cite{wcs} without individual labels to test species disambiguation. This SMALST+WCS dataset covers images of 300 Grevy’s zebras, 232 plains zebras, 150 horses, and 153 donkeys. SMALST is one of the very few fully ID annotated and publicly available datasets for Grevy’s zebra. It covers four images per individual across 37 distinct animals, two-thirds of which are male. Fig.~\ref{fig:pipeline_overview} depicts sample images. Images were captured during the Great Grevy’s Rally~(GGR) in 2018~\cite{grevy_rally_report} over a two-day period at varying times and locations. Note that seeing the right side of a zebra provides no identifiable information on the left side and vice versa. Hence, all pictures taken of the zebras are of the right side. In addition, viewpoints are further labelled as right, front-right, and back-right. The train-test split for the images was generated using stratified sampling, so for each individual there are two images in the train and test set, respectively. To facilitate deep metric learning on this tiny train dataset, we applied  significant offline augmentation, resulting in 50 training images per individual. Augmentations cover random colour jitters (saturation, contrast, and brightness), shifts, and rotation. In addition, we also use 15,000 synthetic images from SMALST~\cite{ZuffiICCV2019} to train 3D fitting.

\begin{figure}[b]
    \centering
    \vspace{-5mm}\includegraphics[width=250pt,height=120pt]{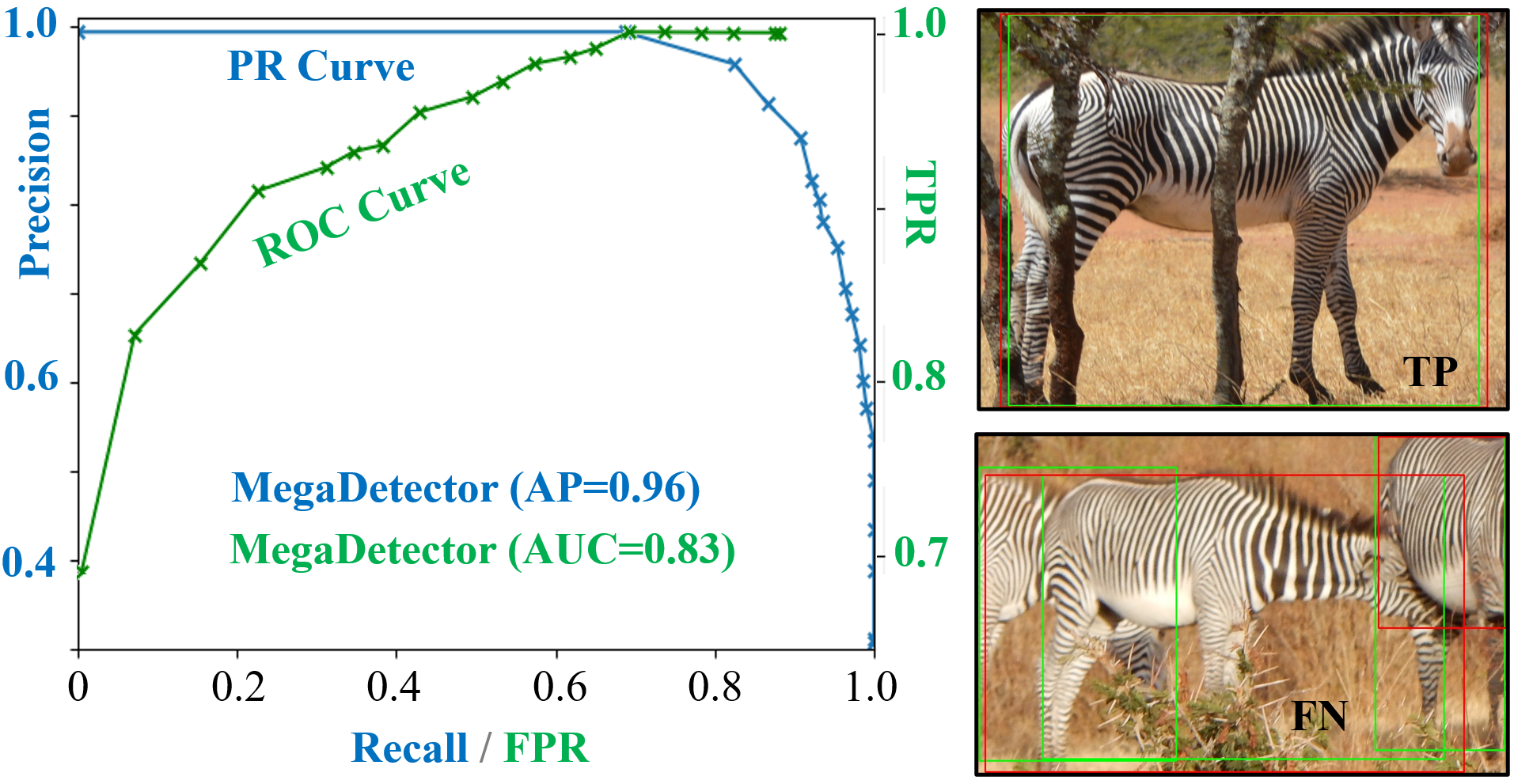}
    \caption{\begin{footnotesize}\textbf{MegaDetector Performance.} \textbf{\textit{(left)}} PR and ROC curve test performance for the MegaDetector on the SMALST dataset. Note the different, colour-coded units and abscissa crossings. \textbf{\textit{(right)}} Example test images from SMALST of a true positive (TP) and a false negative (FN) detection. Whilst partial occlusions by vegetation are generally handled well, animal-animal occlusions often affect the detector's localisation ability negatively. This hints at a link between visual localisation in groups and dazzle coat pattern structure.\end{footnotesize}} \vspace{-10pt}
    \label{fig:mega}
\end{figure}

\section{Methods and Component Performance}

\vspace{-1mm}\subsection{Animal Detection and 2D Localisation}
\begin{figure}[t]

         \centering
         \includegraphics[width=250pt,height=150pt]{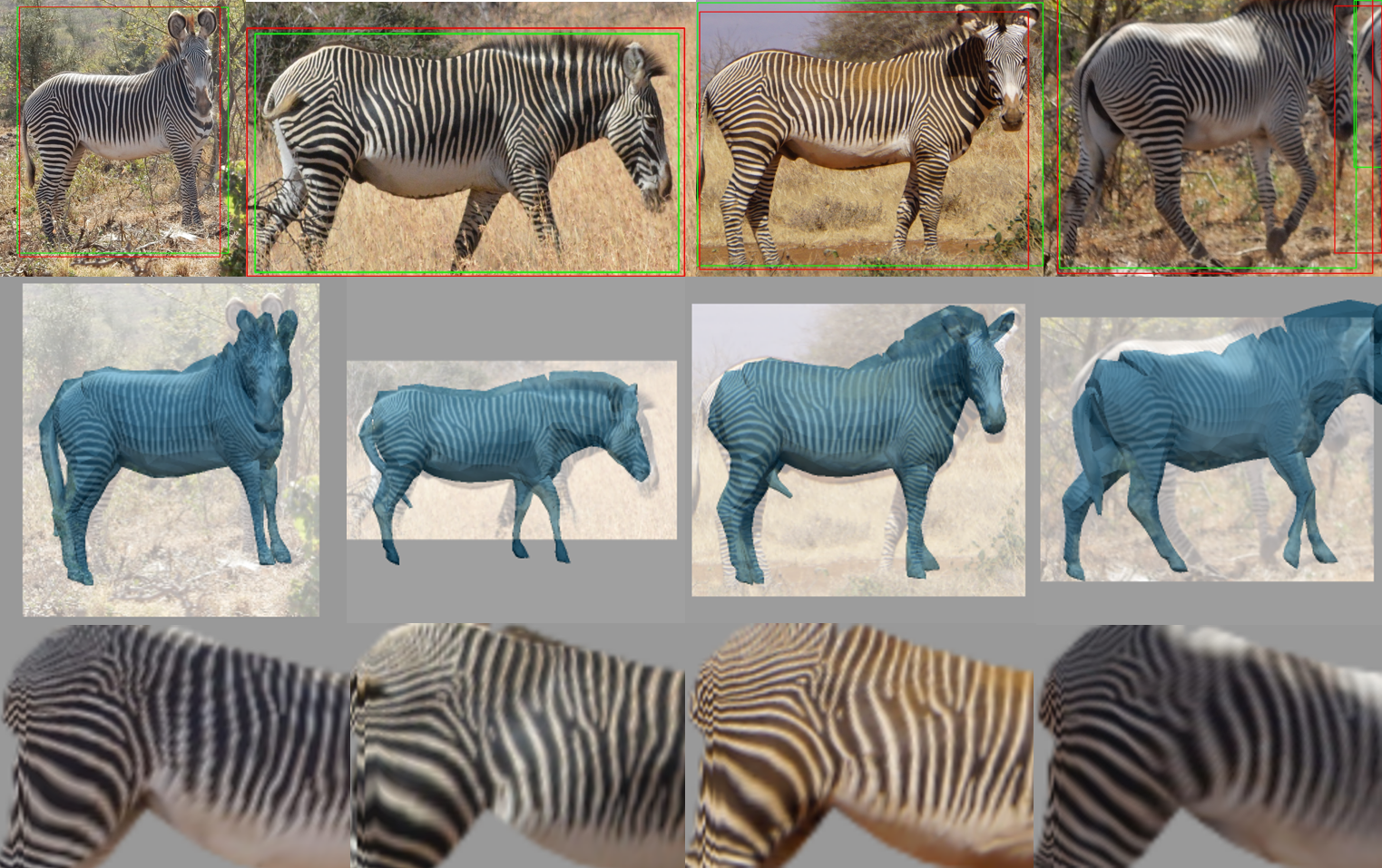}

    \caption{\begin{footnotesize}\textbf{Normalised Coat Pattern Extraction.} The top row shows test image regions containing zebras with the MegaDetector~\cite{MegaDetector} bounding box output (in red) and ground truth (in green). The middle row shows SMALST~\cite{ZuffiICCV2019} fitted 3D zebra models superimposed on the 2D images. The bottom row depicts the back-projected and fully viewpoint and pose normalised texture maps of the zebra hindquarter and back used to identify the individual. Note that the middle two and left/right columns depict the same individual, respectively.\end{footnotesize}}\vspace{-15pt}
    \label{zebra_ind}
\end{figure}

We used v4.1 of the state-of-the-art pre-trained MegaDetector~\cite{MegaDetector} with its F-RCNN ensemble of Inception\cite{https://doi.org/10.48550/arxiv.1409.4842} and ResNet~\cite{https://doi.org/10.48550/arxiv.1512.03385} streams to detect object instances of a generic `animal' class first. We filter out any detections with a confidence score of less than 0.83 (determined via AUC optimisation), or if the detection is nested or nearly inside another (defined by IoU$<0.3$). When evaluated on the SMALST dataset the MegaDetector showed a strong average precision (AP) of 0.96, which allows for region of interest (RoI) generation within the performance range expected for the MegaDetector~\cite{MegaDetector} in different ecosystems. Fig.~\ref{fig:mega}~\textit{(left)} visualises the related performance curves and Fig.~\ref{fig:mega}~\textit{(right)} and Fig.~\ref{zebra_ind}~\textit{(top row)} depict sample bounding box detections.

\subsection{Optional Species Disambiguation}
For applicability to general imagery with visual distractor species, we use the pre-trained species classification model by Microsoft’s AI for Earth team~\cite{sc_model} to perform species disambiguation. This off-the-shelf, non-specific approach fuses Inception and ResNext outputs to confirm Grevy's zebra species presence against any other species. We filter out images where the top species prediction is not of a Grevy’s zebra, irrespective of its species score and summarise results. Using this method on the full SMALST+WCS dataset accuracy was $81.3\%$ at an F1 score of $0.88$. This implies that even in high distractor cases species off-the-shelf disambiguation works fairly well for the species at hand. Note that this step is only required if random wildlife data (e.g. raw camera trap imagery) is fed into the system, whilst most studies apply ID systems to visual archives containing species of interest only.
\subsection{Coat Texture Normalisation Approach}
As bodies are deformable and viewpoints vary, changes in pose or camera angle alter the appearance of coat patterns. By aligning the 2D animal image with a 3D model and back-projecting the coat onto the fitted model surface the pose and viewpoint invariant texture can be estimated explicitly. Fig.~\ref{zebra_ind} depicts test examples of this process. In particular, our pipeline feeds confirmed RoIs of Grevy’s zebra to the SMALST\cite{ZuffiICCV2019} regression framework using SMAL horse templates~\cite{Zuffi:CVPR:2017,Zuffi:CVPR:2018} to estimate the 3D shape, pose, and ultimately visible coat pattern surface texture.

\subsection{Deep 3D Fitting Network}
We follow the approach by Zuffi \textit{et al.}~\cite{ZuffiICCV2019} for training the texture fitting network. It consists of a ResNet backbone followed by a convolutional layer and 2 fully connected layers which produce a 1024-dimensional feature vector summarising appearance information of the input. This feature vector is fed to independent subsequent layers that predict shape, 3D pose, and texture maps. Shape prediction is implemented via fully connected layers that produce a 40-dimensional shape feature, used to predict vertex deformations applied to the SMAL 3D model. Pose prediction is performed by a linear layer that outputs a vector of relative joint angles. This vector sets the 3D pose of the SMAL model complemented by camera frame translation prediction via two linear layers that output coordinates (x,y,z). Finally, texture prediction uses encoder-decoder prediction of a uv-flow map from four stitched sub-images~\cite{Zuffi:CVPR:2017}. We extract a compact subset from this map containing hindquarter and back areas to cover key unique portions of the coat (see Fig.~\ref{zebra_ind}~\textit{(bottom row)}). Other characteristic areas such as scapular and neck regions could be added in further studies, but suffer from (self)occlusion and shadows more frequently.  

\begin{figure}[b]
    \centering
    \includegraphics[width=250pt]{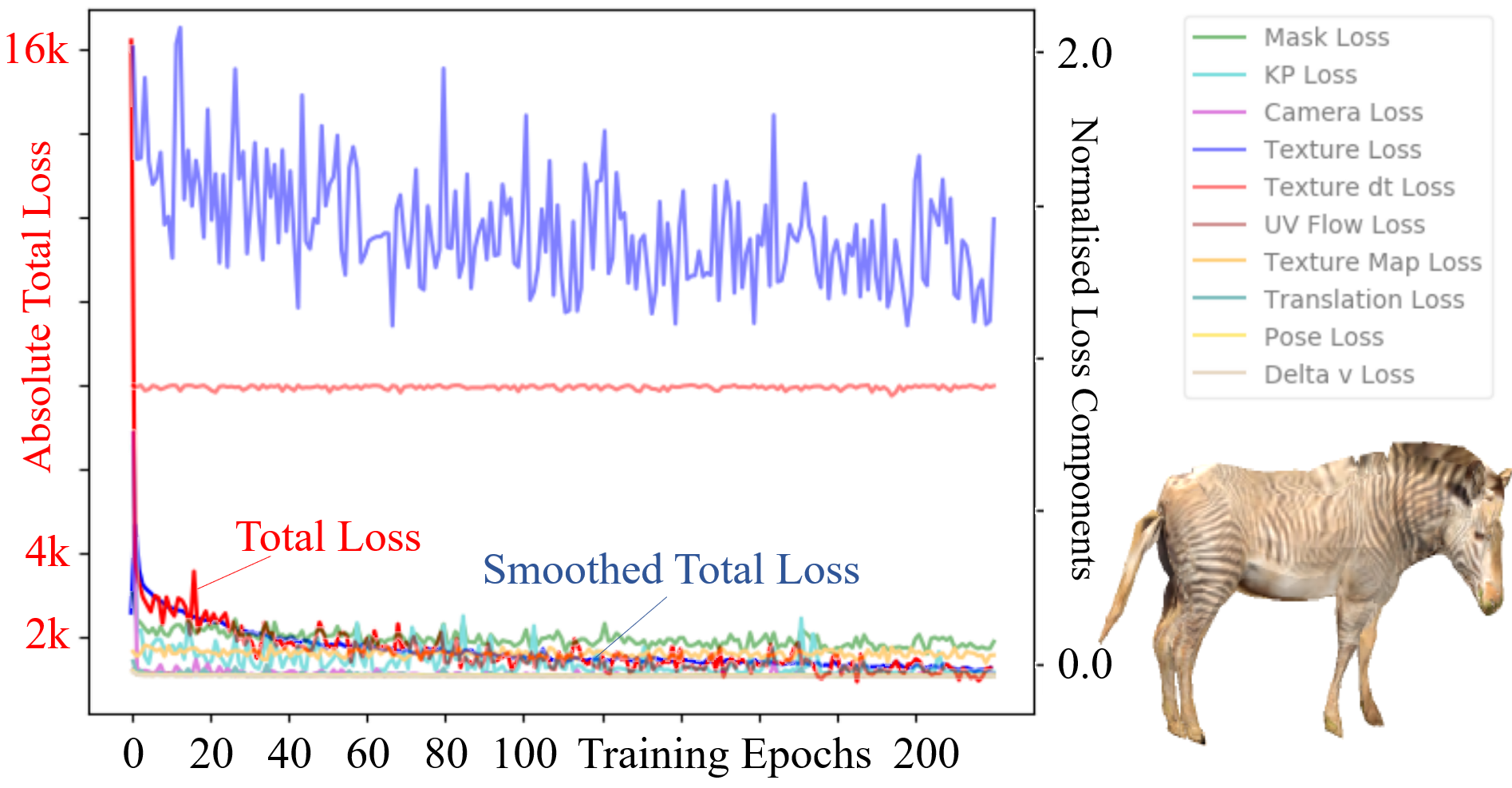}
    \caption{\begin{footnotesize}\textbf{Deep 3D Model Training.} The graph depicts loss evolution during network training for 3D model fitting. Note that most losses components drop very quickly apart from the texture loss (top curve in light blue), which only gradually reduces with learning. Loss components on the right are explained in detail in~\cite{ZuffiICCV2019}. An example of a fitted and textured 3D model is also shown.\end{footnotesize}} \vspace{-10pt}
    \label{fig:fitting}
\end{figure}

\subsection{Individual Identification via Metric Learning}
We map cropped textures into learned, individually distinctive latent space following Andrew \textit{et al.}\cite{andrew2020visual,lagunes2019learning}. As a result, texture mappings of the same individual naturally cluster together. The space is built via triplets of inputs, which include an anchor pattern, a positive (pattern of the same individual as the anchor), and a negative (different individual to the anchor). A ResNet model optimised via SoftMax and reciprocal triplet losses is used to learn the visual similarities and dissimilarities between the individuals forming an identity space~(see Fig.~\ref{embedding} for a space visualisation). Unseen input textures can then be projected into this domain and a k-Nearest Neighbours approach will reveal closest individual identities.

\section{ID Experiments \& Results}
\begin{table}[t]
\begin{center}
 \includegraphics[width=250pt]{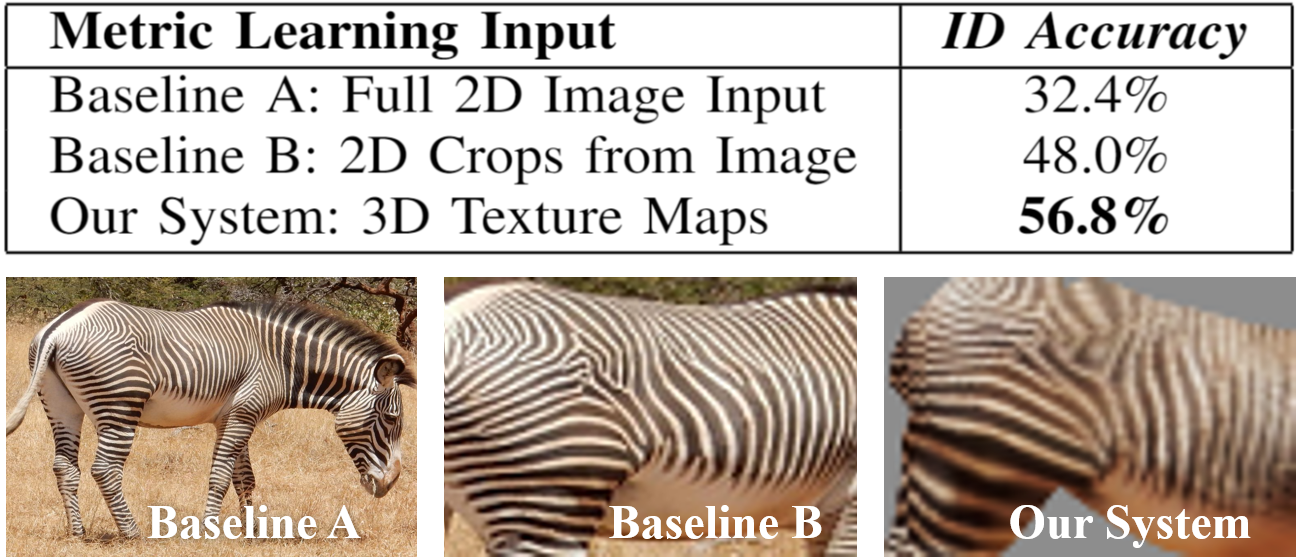}\vspace{-5pt}
\end{center}\caption{\begin{footnotesize}\textbf{2D vs 3D Identification Performance.} Top-1 accuracy results for three metric learning setups for 74 test images of 37 individuals of the SMALST dataset differentiating our 3D pipeline against two 2D baselines. Input examples are given at the bottom.\end{footnotesize}}\vspace{-11pt}
\label{results_table}
\end{table}
\vspace{-1mm}\subsection{3D Fitting Training}
We trained the 3D fitting network using 15,000 synthetic images from the SMALST training portion. Training proceeded for 210 epochs at a learning rate of 0.0001 and batch size 8 with a texture resolution of 256x256. Fig.~\ref{fig:fitting} illustrates this training process quantitatively. We retained the network that performed best on the SMALST validation set in relation to the Percentage of Correct Keypoints (PCK)@0.1 and used it to extract normalised texture maps. Training ran for approx. 8 days on a BlueCrystal NVIDIA P100 node utilising one GPU.

\subsection{Metric Space Training}
Half of the images from SMALST are used for metric learning where each image is 3$\times$ augmented increasing the training set size by factor 25. Further augmentations were not performed as it resulted in a plateau in performance. An online triplet mining method~\cite{Schroff_2015} known as `batch hard' was used for sampling from this training set. We trained the network for 200 epochs employing a batch size of 8, a learning rate of 0.1, and a reciprocal triplet lambda~\cite{andrew2020visual,lagunes2019learning} of 0.0001. Training took approx. 4 hours utilising the hardware setup detailed above.

\subsection{Basic Identification Experiments}
\begin{figure}[t]
    \centering
    \vspace{-10pt}\includegraphics[width=9cm,height=5.2cm]{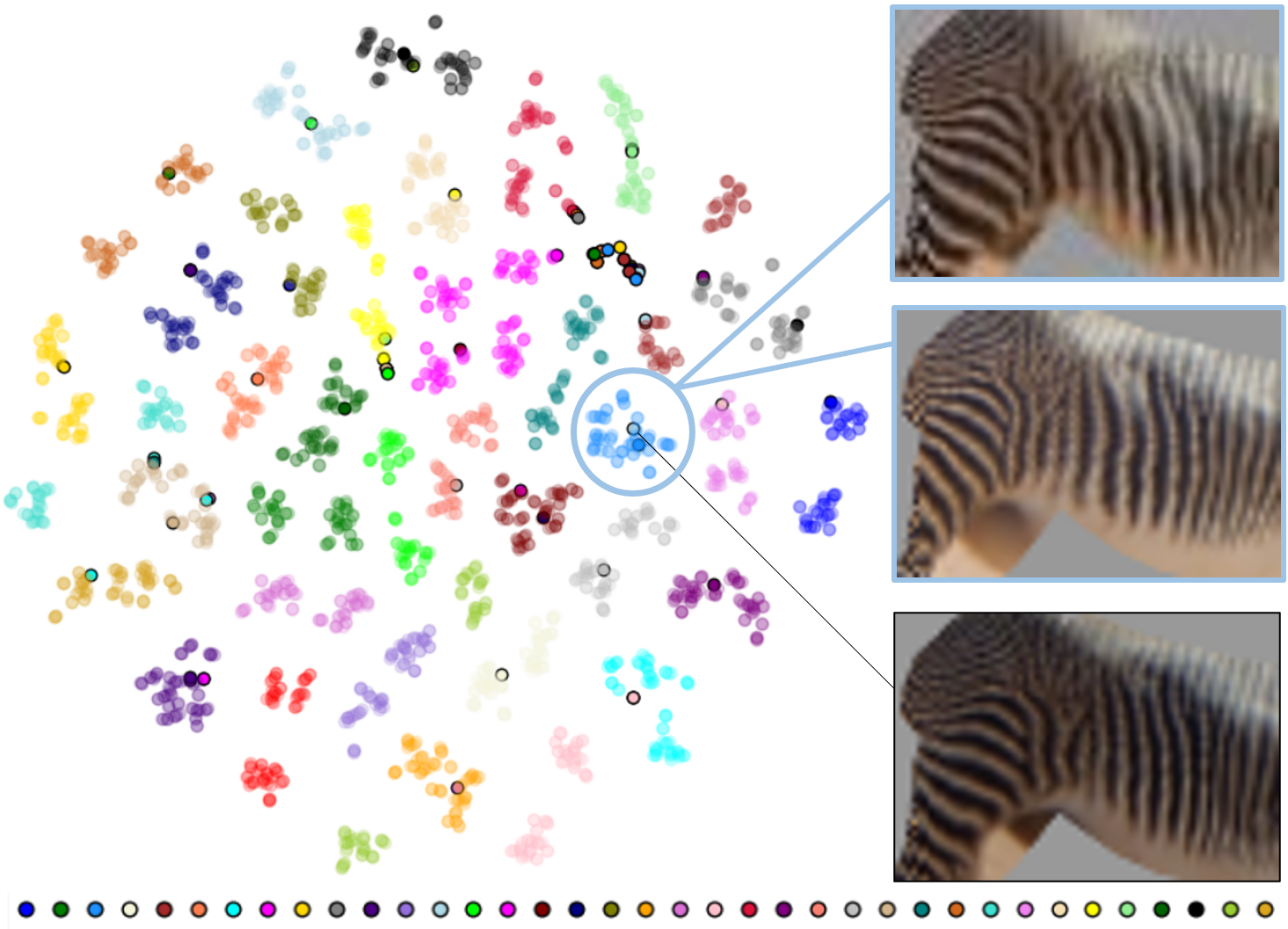}
    \caption{\begin{footnotesize}\textbf{Metric Identity Space of SMALST Population.} We depict a 2D embedding of the 128D identity space generated by our pipeline covering the SMALST Grevy's zebra population containing 37 individuals. We use t-distributed Stochastic Neighbour Embedding (t-SNE) for dimensionality reduction. The colour code at the bottom maps to individual IDs left to right. Test data points carry a black outline, training data points are shown as simple colour discs. A sample pattern from the test set (bottom right) and two training patterns for a particular individual are also depicted. Note the very clean clustering of training information in this space. However, due to the tiny dataset the system overfits and various test data points are not reliably identified via the kNN approach. Future work will address these limitations.\end{footnotesize}}\vspace{-15pt}
    \label{embedding}
\end{figure}

We tested our pipeline against two simple baselines using SMALST data. First, we fed whole images without animal localisation into the metric learning component. Secondly, we used 2D texture patches of the zebra hindquarters and backs cropped from the 2D image by an F-RCNN localiser to assess in how far 2D data can compete with our 3D design. Results of our system versus these two baselines are shown in Table~\ref{results_table} and Fig.~\ref{embedding} visualises our identity space.

\section{Prospects and Conclusion}\vspace{-5pt}
This paper combined deep learning techniques for species detection, 3D model fitting, and metric learning to perform individual animal ID from photographs. This is the first work to attempt deep 3D ID. Whilst 2D computer vision algorithms changed the scale and scope individual-based ecological analyses, these techniques often require 2D images be filtered by pose, orientation, and clarity. Expanding image analysis to 3D provides significant advantages since fitting images to mannequin-like 3D poses reduces coat texture deformation, adds explicit pose information to the output, and can aid algorithmic accuracy as shown in this preliminary study. We applied our 3D approach to individual Grevy's zebra \textit{(Equus grevyi)}. We showed on the SMALST test dataset that the use of 3D model fitting can indeed benefit performance. However, the study is too small accurately to suppress overfitting and estimate the full performance potential achievable in larger-scale real-world application. Large training data sets, higher resolution texture maps, and integration with existing non-deep frameworks will be needed to judge the pipeline w.r.t. state-of-the-art ID performance~\cite{hotspotter,andrew2020visual,schneider}. However, being able to construct and visualise population spaces based on pose-normalised animal coat data and associate pose and ID information to this can benefit a range of behavioral and ecological studies. Further collaboration between the computer science and biological communities will be essential to enhance the data and algorithm base to improve on the quality of the biology that relies on re-sighting individuals.

\vspace{5pt}
\begin{scriptsize}
\noindent\textbf{Acknowledgements:} We appreciate publications by GGR/SMALST, WCS, MegaDetector, AI4Earth, and Andrew/Lagunes. Thanks to T Berger-Wolf, C Stewart, and J Parham. This work was carried out using the computational facilities of the Advanced Computing Research Centre, University of Bristol - http://www.bris.ac.uk/acrc/.
\end{scriptsize}






%
\bibliographystyle{IEEEtran}
\vspace{-10pt}
\begin{scriptsize}
\bibliography{grevy}

\begin{thebibliography}{10}
\providecommand{\url}[1]{#1}
\csname url@samestyle\endcsname
\providecommand{\newblock}{\relax}
\providecommand{\bibinfo}[2]{#2}
\providecommand{\BIBentrySTDinterwordspacing}{\spaceskip=0pt\relax}
\providecommand{\BIBentryALTinterwordstretchfactor}{4}
\providecommand{\BIBentryALTinterwordspacing}{\spaceskip=\fontdimen2\font plus
\BIBentryALTinterwordstretchfactor\fontdimen3\font minus
  \fontdimen4\font\relax}
\providecommand{\BIBforeignlanguage}[2]{{%
\expandafter\ifx\csname l@#1\endcsname\relax
\typeout{** WARNING: IEEEtran.bst: No hyphenation pattern has been}%
\typeout{** loaded for the language `#1'. Using the pattern for}%
\typeout{** the default language instead.}%
\else
\language=\csname l@#1\endcsname
\fi
#2}}
\providecommand{\BIBdecl}{\relax}
\BIBdecl

\bibitem{MegaDetector}
S.~Beery, D.~Morris, and S.~Yang, ``Efficient pipeline for camera trap image
  review,'' in \emph{Data Mining and AI for Conservation at KDD}, 2019.

\bibitem{sc_model}
\BIBentryALTinterwordspacing
Microsoft, ``Species classification github repository,'' {Accessed:
  2022-06-24}. [Online]. Available:
  \url{https://github.com/microsoft/SpeciesClassification}
\BIBentrySTDinterwordspacing

\bibitem{ZuffiICCV2019}
S.~Zuffi, A.~Kanazawa, T.~Berger-Wolf, and M.~J. Black, ``{Three-D Safari:
  Learning to Estimate Zebra Pose, Shape, and Texture from Images In the
  Wild},'' in \emph{ICCV}, Oct. 2019.

\bibitem{andrew2020visual}
W.~Andrew, J.~Gao, S.~Mullan, N.~Campbell, A.~Dowsey, and T.~Burghardt,
  ``Visual identification of individual {Holstein-Friesian} cattle via deep
  metric learning,'' \emph{Computers and Electronics in Agriculture}, vol. 185,
  2021.

\bibitem{tuia}
D.~Tuia, B.~Kellenberger, S.~Beery, B.~R. Costelloe, S.~Zuffi, B.~Risse,
  A.~Mathis, M.~W. Mathis, F.~van Langevelde, T.~Burghardt \emph{et~al.},
  ``Perspectives in machine learning for wildlife conservation,'' \emph{Nature
  communications}, vol.~13, no.~1, pp. 1--15, 2022.

\bibitem{petersen}
J.~Petersen, ``{An identification system for zebra (Equus burchelli, gray)},''
  \emph{African Journal of Ecology}, no.~10, pp. 69--63, 1972.

\bibitem{hiby}
L.~Hiby and P.~Lovell, ``Computer aided matching of natural markings: a
  prototype system for grey seals,'' \emph{Report of the International Whaling
  Commission}, no.~12, pp. 57--61, 1990.

\bibitem{lahiri}
M.~Lahiri, C.~Tantipathananandh, R.~Warungu, D.~I. Rubenstein, and T.~Y.
  Berger-Wolf, ``Biometric animal databases from field photographs:
  identification of individual zebra in the wild,'' in \emph{ACM International
  Conference on Multimedia Retrieval}, 2011, pp. 1--8.

\bibitem{sherley}
R.~B. Sherley, T.~Burghardt, P.~J. Barham, N.~Campbell, and I.~C. Cuthill,
  ``Spotting the difference: towards fully-automated population monitoring of
  african penguins spheniscus demersus,'' \emph{Endangered Species Research},
  vol.~11, no.~2, pp. 101--111, 2010.

\bibitem{hotspotter}
J.~Crall, C.~Stewart, T.~Berger-Wolf, D.~Rubenstein, and S.~Sundaresan,
  ``Hotspotter - patterned species instance recognition,'' in \emph{2013 IEEE
  Workshop on Applications of Computer Vision}, 2013, pp. 230--237.

\bibitem{parham}
J.~Parham, C.~Stewart, J.~Crall, D.~Rubenstein, J.~Holmberg, and
  T.~Berger-Wolf, ``An animal detection pipeline for identification,'' in
  \emph{WACV}, 2018, pp. 1075--1083.

\bibitem{berger}
T.~Y. Berger-Wolf, D.~I. Rubenstein, C.~V. Stewart, J.~A. Holmberg, J.~Parham,
  S.~Menon, J.~Crall, J.~V. Oast, E.~Kiciman, and L.~Joppa, ``Wildbook:
  Crowdsourcing, computer vision, and data science for conservation,'' in
  \emph{Bloomberg Data for Good Exchange Conference}, 2019.

\bibitem{schneider}
S.~Schneider, G.~W. Taylor, and S.~C. Kremer, ``Similarity learning networks
  for animal individual re-identification-beyond the capabilities of a human
  observer,'' in \emph{Proceedings of the IEEE/CVF Winter Conf. on Applications
  of Computer Vision Workshops}, 2020, pp. 44--52.

\bibitem{wcs}
\BIBentryALTinterwordspacing
W.~C. Society, ``Wildlife conservation society {(WCS)} camera traps,''
  {Accessed: 2022-06-30}. [Online]. Available:
  \url{https://lila.science/datasets/wcscameratraps}
\BIBentrySTDinterwordspacing

\bibitem{grevy_rally_report}
\BIBentryALTinterwordspacing
D.~Rubenstein, J.~Parham, C.~Stewart, T.~Y. Berger-Wolf, J.~Holmberg, J.~Crall,
  B.~L. Mackey, S.~Funnel, K.~Cockerill, Z.~Davidson, L.~Mate, C.~Nzomo,
  R.~Warungu, D.~Martins, V.~Ontita, J.~Omulupi, J.~Weston, G.~Anyona,
  G.~Chege, D.~Kimiti, K.~Tombak, A.~Gersick, and N.~Rubenstein, ``{The State
  of Kenya’s Grevy’s Zebras and Reticulated Giraffes: Results of the Great
  Grevy’s Rally 2018},'' {Accessed: 2022-06-30}. [Online]. Available:
  \url{https://wildlifedirect.org/wp-content/uploads/2018/06/CS-report-GGR-2018v-4.pdf}
\BIBentrySTDinterwordspacing

\bibitem{https://doi.org/10.48550/arxiv.1409.4842}
C.~Szegedy, W.~Liu, Y.~Jia, P.~Sermanet, S.~Reed, D.~Anguelov, D.~Erhan,
  V.~Vanhoucke, and A.~Rabinovich, ``Going deeper with convolutions,'' in
  \emph{CVPR}, 2015, pp. 1--9.

\bibitem{https://doi.org/10.48550/arxiv.1512.03385}
K.~He, X.~Zhang, S.~Ren, and J.~Sun, ``Deep residual learning for image
  recognition,'' in \emph{2016 IEEE Conf. on Computer Vision and Pattern
  Recognition}, 2016, pp. 770--778.

\bibitem{Zuffi:CVPR:2017}
S.~Zuffi, A.~Kanazawa, D.~Jacobs, and M.~J. Black, ``{3D} menagerie: Modeling
  the {3D} shape and pose of animals,'' in \emph{IEEE Conf. on Computer Vision
  and Pattern Recognition}, Jul. 2017.

\bibitem{Zuffi:CVPR:2018}
S.~Zuffi, A.~Kanazawa, and M.~J. Black, ``Lions and tigers and bears: Capturing
  non-rigid, {3D}, articulated shape from images,'' in \emph{CVPR}, 2018.

\bibitem{lagunes2019learning}
M.~Lagunes-Fortiz, D.~Damen, and W.~Mayol-Cuevas, ``Learning discriminative
  embeddings for object recognition on-the-fly,'' in \emph{2019 International
  Conf. on Robotics and Automation}.\hskip 1em plus 0.5em minus 0.4em\relax
  IEEE, 2019, pp. 2932--2938.

\bibitem{Schroff_2015}
F.~Schroff, D.~Kalenichenko, and J.~Philbin, ``{FaceNet}: A unified embedding
  for face recognition and clustering,'' in \emph{2015 {IEEE} Conf. on Computer
  Vision and Pattern Recognition ({CVPR})}.\hskip 1em plus 0.5em minus
  0.4em\relax {IEEE}, jun 2015.

\end{thebibliography}
\end{scriptsize}

\end{document}